# Modelling Resistive and Phase Change Memory with Passive Selector Arrays – A Matlab Tool

Yasir J. Noori, *IEEE member* and C. H. (Kees) De Groot, *IEEE Senior Member*

*Abstract*—Memristor devices are crucial for developing neuromorphic computers and next-generation memory technologies. In this work, we provide a comprehensive modelling tool for simulating static DC reading operations of memristor crossbar arrays that use passive selectors with matrix algebra in MATLAB. The software tool was parallel coded and optimized to run with personal computers and distributed computer clusters with minimized CPU and memory consumption. Using the tool, we demonstrate the effect of changing the line resistance, array size, voltage selection scheme, selector diode's ideality factor, reverse saturation current, temperature and sense resistance on the electrical behavior and expected sense margin of one-diode-one-resistor crossbar arrays. We then investigate the effect of single and dual side array biasing and grounding on the dissipated current throughout the array cells. The tool we offer to the memristor community and the studies we present enables the design of larger and more practical memristor arrays for application in data storage and neuromorphic computing.

## I. INTRODUCTION

MEMRISTOR hardware is being extensively developed as artificial synapses, inspired by the brain intelligence and the efficient information processing it is capable of [1]–[3]. This has the potential to achieve major breakthroughs in pattern recognition and machine learning. In addition, CMOS based resistive RAMs (ReRAM) and non-volatile phase-change memories (PCM) are being developed by major industrial players, such as Intel and Micron Technology, for applications in the memory-storage space, motivated by their scalable device footprint and high switching speed [4]–[7]. The roadmap of phase change memories anticipates the technology to bridge the gap between the fast but low bit density dynamic random access memory (DRAM) and the slow but relatively higher bit density flash technology in a hybrid memory system [8]–[10].

Memristor architectures have been primarily based around the simple crossbar array structure. The simplicity of crossbar arrays can allow the realization of high device density in two and three-dimensions whilst enabling low fabrication and production costs [11]–[14]. However, designing memristive crossbar arrays require rigorous quantitative electrical analysis of the system to assess its performance. While there have been considerable efforts to model crossbar arrays in the past, in most attempts, the selector device parameters and line resistances were not included in the models. Most of these crossbar array modellings have been done using the SPICE modelling tool [15]–[17]. However, modelling large memory arrays above a Megabit requires extensive computational power with SPICE [18]. Although SPICE is a compact tool that is highly optimized for modelling complicated electronic circuits, the nature of node analysis makes it slow for modelling very large memory arrays. Therefore, a highly parallelized MATLAB tool that can perform the array simulation with matrix algebra utilizing large supercomputer clusters, makes modelling future high-density memory arrays much more practical for research and commercial purposes [19].

There are two aspects to the novelty of this work. Firstly, we are providing a parallelized open-access software tool for the memristor scientific community that can be used to model memristor crossbar arrays with passive selector devices. This work follows from the theoretical work of An Chen which proposes a comprehensive crossbar array model that incorporates both line resistance and nonlinear device characteristics [20]. Secondly, we extend his work by utilizing the Lambert-W function for simulating reading operations of diode-memristor crossbar arrays. The function allows incorporating the selector diode's ideality factor, reverse saturation current and temperature as simulatable parameters in the algorithm of the tool. Compared to previous works, this is the first work that shows a simulation of a comprehensive list of all the input parameters of an array, particularly focusing on optimizing its performance for different selector parameters under different read schemes.

The code of the tool we provide was made to run on supercomputer clusters utilizing the MATLAB Distributed Computing Server toolbox. The code is optimized to reduce memory and CPU usage to allow the simulation of many megabit memory arrays in a time that is orders of magnitude shorter than what SPICE can achieve. In order to evaluate the performance of the tool, we present a quick overview of the background theory and method of the modelling tool and propose different simulation scenarios and the results that it can output for each in one diode one memristor arrays (1D1R) setup. We also present more details related to programming the tool and optimizing its efficiency in section IV.

## II. METHOD

A $m \times n$ crossbar array such as the one shown in Fig. 1, can

Manuscript received June 20, 2019. This work was funded by the EPSRC programme grant ADEPT – Advanced Devices by ElectroPlaTing, EPSRC reference: EP/N035437/1. The authors would like to thank Ruomeng Huang for suggesting this research direction.

The authors are affiliated with the school of Electronics and Computer Science, The University of Southampton, Southampton, SO17 1BJ, UK.

(Corresponding author: Yasir J. Noori, y.j.noori@southampton.ac.uk).

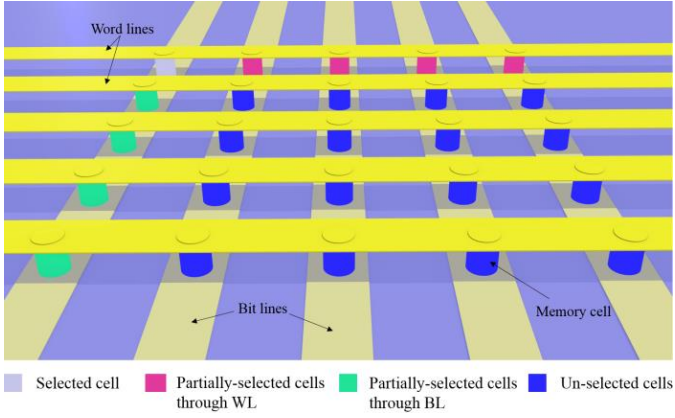

Fig. 1. An illustration of the structure of a typical crossbar resistive memory array showing the selected, partially selected and unselected cells.

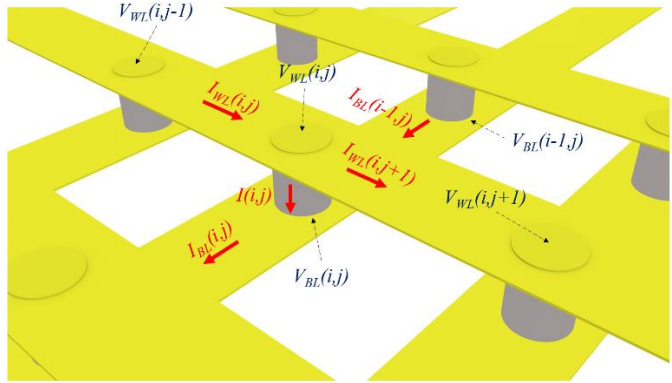

Fig. 2. An illustrative diagram for Eq.1 showing the flow of current through a junction in a crossbar memory array.

be modelled using Kirchhoff's current equation at every junction point, as shown in Fig. 2. There are two equations that model the current flow through the corresponding word line (WL) and bit line (BL) at every junction

$$I_{WL}(i,j) = I(i,j) + I_{WL}(i, j + 1)$$
$$I_{BL}(i,j) = I(i,j) + I_{BL}(i - 1, j) \quad (1)$$

These can be written in terms of the voltages at each junction for WL and BL. This produces six equations, four of which relate to cells at the edges of the array where their voltages correspond to the applied voltage at both sides of the WL ($V_{App\_WL1}$ and $V_{App\_WL2}$) and BL ($V_{App\_BL1}$ and $V_{App\_BL2}$). These equations can be written in matrix form using MATLAB. We refer the reader to the appendix of ref [20] for a fully detailed mathematical listing of the equations and the simulation algorithm.

Fig. 3 illustrates graphically the input parameters that are required for the code to solve the voltages and currents in the circuit. The output parameters that can be found using this tool include the array's junction currents $I(i,j)$, $I_{WL}(i,j)$, $I_{BL}(i,j)$, junction voltages $V_{WL}(i,j)$, $V_{BL}(i,j)$, making it possible to calculate leakage current, cell power dissipation and sense margin. The tool can also be used to visually represent the variation of those values throughout a crossbar array to demonstrate the effects of the input parameters, as will be shown later.

Rectifying diodes are strong selector candidates for crossbar resistive switching memory systems. Modelling a selector diode in series with a memristor device requires solving the non-linear equation that arises due to the selector. To model a 1D1R crossbar array we propose using the Lambert-W function (also known as the Omega function). The cell schematic modelled in this work is a selector diode connected in series with a constant resistor representing a high or low resistance memristive state, as shown in Fig. 4.

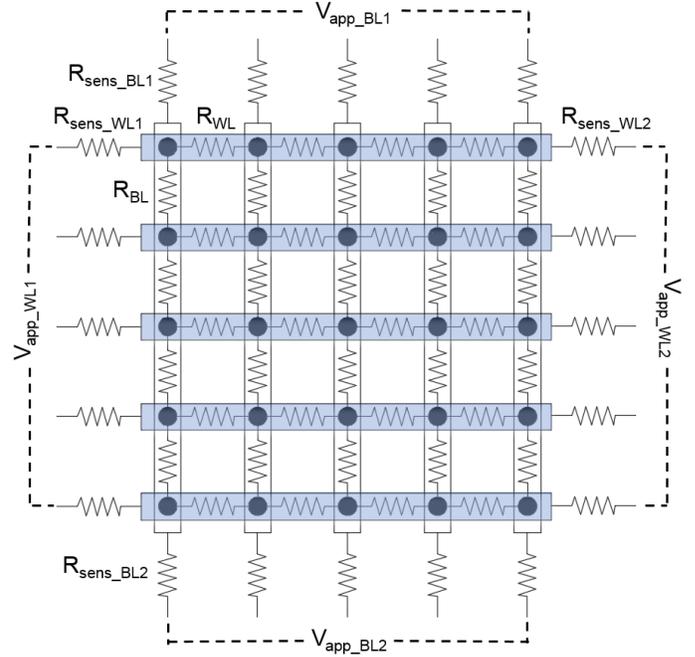

Fig. 3. An illustration diagram of the input parameters for the proposed crossbar array simulator. In addition to the above input parameters, the selector's parameters include the ideality factor (η), reverse saturation current ($I_s$), temperature (T) and reverse bias current (Rs)

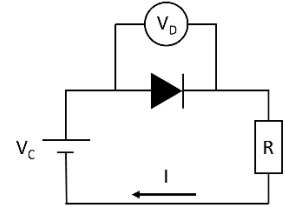

Fig. 4. A schematic diagram representing a model of a selector diode in series with a resistor.

The Lambert-W function can be used to model a diode-resistor circuit as shown in Eq. 2

$$I \approx \frac{\eta V_T}{R} W\left(\frac{I_s R}{\eta V_T} e^{\frac{V_C}{\eta V_T}}\right) \quad (2)$$

where the thermal voltage $V_T = \frac{k_B T}{q}$, $V_C$ is the potential difference across a cell, $I_s$ is the reverse saturation current, $\eta$ is the selector's ideality factor, $R$ is the resistance of the memristor device, $k_B$ is Boltzmann's constant, $T$ is the device temperature and $q$ is the electron's charge. This can be derived from the original diode formula (Eq. 3) as shown in reference [21].

$$I = I_s(e^{\frac{V_D}{\eta V_T}} - 1) \quad (3)$$

The advantages of using the Lambert-W function is that it is pre-implemented and optimized in MATLAB, making it an ideal method to be used in this tool. The tool works by initially setting $V_C$ to be equivalent to the applied voltage, $V_{app\_WL}$. The equivalent resistances of the cells' resistor and diode are calculated through the differential of Eq. 2 with respect to $V_C$. The new array resistances are then feedbacked to the code and a new set of array WL and BL voltages are calculated. The equivalent resistance of every cell is then calculated using the new set of voltages. This process is repeated until a solution is found where the change in node voltages between two iterations is within an acceptable error, i.e. the solution must converge.

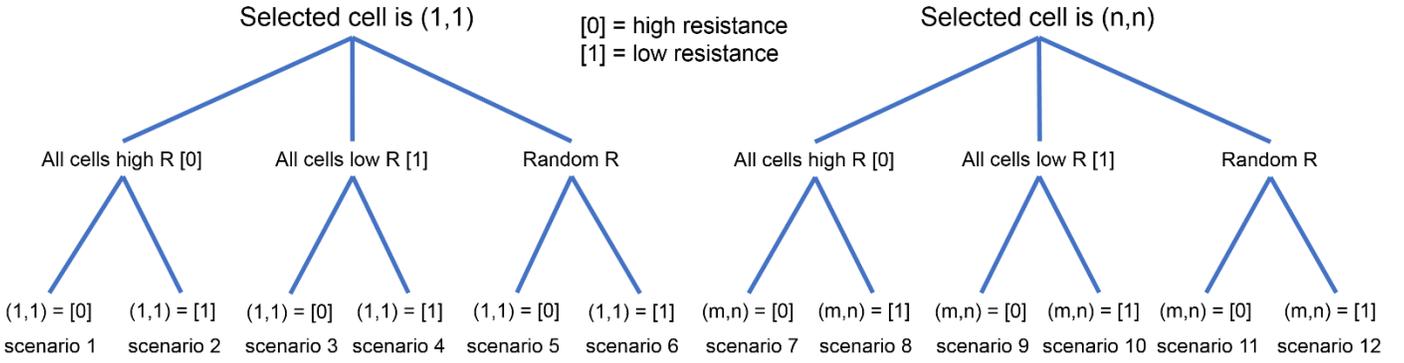

Fig. 5. An illustration diagram showing the proposed 12 state scenarios of a memory array for evaluating the tool.

Simulating large memory arrays is repetitive, in the sense that it involves a great number of loops that scales dramatically as the simulated array's size is increased. For-loops were extensively used in this tool for building the input matrices (matrices A, B, C, D and E from the appendix of ref [20]) of the simulation and finding the voltage solution. To optimize the speed of the process, the tool was coded to utilize multi-core processors using for-loops of the type "*parfor*". The *parfor*-loop type can be utilized by installing the "MATLAB Parallel Computing Toolbox". In this tool, *parfor*-loops were found to use 100% of all available processer's time when executed, demonstrating the efficiency of parallel processing in this tool which makes it ideal for use in supercomputers. In addition, storing all the data generated during the simulations requires an extensive amount of memory. To overcome this, the data was stored using matrices of the type "*sparse*" to ensure the storage memory is not wasted in storing zero matrix elements. This allowed reducing the memory requirements from over 40 GB for 40×40 arrays to 2 GB for 1000×1000 arrays.

To evaluate the results obtained from our tool and to assess its validity, the effect of line resistance, array size and voltage selection scheme on the apparent resistance were tested quantitatively. These are particularly important in designing dense and large crossbar memory arrays. We designed a tool testing protocol that characterizes the electrical performance of the array using several possible memory state scenarios. The results from these scenarios should give a general qualitative overview of the expected results from any other possible array state, including best- and worst-case scenarios, the protocol is illustrated in Fig. 5. In our testing protocol, there are a total of twelve scenarios that can be run in parallel and produce apparent resistance values. In the twelve scenarios, six of those are made to measure the resistance of the cells located at the closest corner to the voltage source and ground $(1,1)$, while the other six scenarios are made to measure that for the cell located at the furthest corner $(m, n)$. Both scenario groups are dedicated to measuring the apparent resistance of the selected cell $(R_{select})$ when it is high and low, while the unselected cells $(R_{unselect})$ are at high, low and random states.

The apparent resistance of $R_{select}$ is calculated by dividing the voltage difference between the applied potential at the selected WL and voltage at the sense resistor $(R_{sens})$ by the total current collected through selected BL's end

$$R = \left(\frac{V_{app\_WL} - V_{sens\_BL}}{\sum_{k=0}^{m} I(k, j_R)}\right) \quad (4)$$

The key figure of merit that needs to be considered in the design of selector devices is the sense margin, which determines the smallest sense voltage window for reading operations. This is identified as the percentage difference between the sense voltages for a cell's low and high resistance states normalized to the input voltage.

In a $m \times n$ crossbar array, there can be $2^{m \times n}$ distinct digital states for the matrix. In a read operation, the worst-case scenario is defined as the scenario when the sense margin is minimum. This occurs when the voltage drop across $R_{sens}$ is the smallest while $R_{select} = R_{low}$ and the voltage drop is the largest when the $R_{select} = R_{high}$. The worst-case scenario involves selecting the cell that is located at the furthest corner from the voltage source and ground, due to the finite WL and BL resistance. In our simulations, the sense margin is calculated using the change in voltage dropped across the sense resistor between scenarios 8 and 9 [16], [22]. This should not be confused with the worst-case scenario for write operations which involves selecting the cell at the furthest corner from the voltage source and ground, while all $R_{unselect} = R_{low}$. Hence, in our model, the sense margin is calculated as:

$$sense\ margin(\%) = \frac{(V_{sens\_BL\_sc8}(j) - V_{sens\_BL\_sc9}(j)) \times 100}{V_{app\_WL}(i)} \quad (5)$$

To ensure that the software tool is working as expected, we performed several tests to calculate the apparent resistance of $R_{select}$ from the 12 scenarios described previously. The results for a 100×100 array that contains one resistor and one diode component in every cell is shown in Fig. 6. The calculated apparent resistance values for scenarios 1,3,5,7,9 and 11 are plotted in red, representing that in these scenarios $R_{select} = R_{high}$. On the other hand, the apparent values for scenarios 2, 4, 6, 8, 10 and 12 are plotted in blue, representing $R_{select} = R_{low}$. The input parameters here were chosen deliberately to obtain a sparse range of calculated resistances for results evaluation. Unless stated otherwise, table I lists all the input parameters for all the simulations that were done in this work.

In all the next simulations, the line resistance for WL $(R_{WL})$ and BL $(R_{BL})$ were chosen to be the same for simplicity and will be symbolized as $R_l$, but they can be independently adjusted in the tool. All the simulations were done using a square array for simplicity, however, arbitrary shaped 2D arrays can also be simulated. The iterative simulation process for these results was stopped after reaching an error smaller than 0.01%. Increasing the accuracy by a factor of 100 will approximately double the simulation time.

| Input parameters | Value | Input parameters | Value |
| --- | --- | --- | --- |
| $R_{low}$ | $10\ k\Omega$ | $V_{app\_WL1\_selected}$ | $1\ V$ |
| $R_{high}$ | $1\ M\Omega$ | $V_{app\_WL1\_unselected}$ | $0.5\ V$ |
| $R_{sens\_WL1}$ | $10\ \Omega$ | $V_{app\_WL2\_selected}$ | $0\ V$ |
| $R_{sens\_WL2}$ | $100\ M\Omega$ | $V_{app\_WL2\_unselected}$ | $0\ V$ |
| $R_{sens\_BL1\_unselected}$ | $10\ \Omega$ | $V_{app\_BL1\_selected}$ | $0\ V$ |
| $R_{sens\_BL1\_selected}$ | $1\ k\Omega$ | $V_{app\_BL1\_unselected}$ | $0.5\ V$ |
| $R_{sens\_BL2\_unselected}$ | $100\ M\Omega$ | $V_{app\_BL2\_selected}$ | $0\ V$ |
| $R_{sens\_BL2\_selected}$ | $100\ M\Omega$ | $V_{app\_BL2\_unselected}$ | $0\ V$ |
| Diode Temperature (T) | $300\ K$ | | |

Table I: The input parameters that were chosen for the simulations results that are presented in this paper.

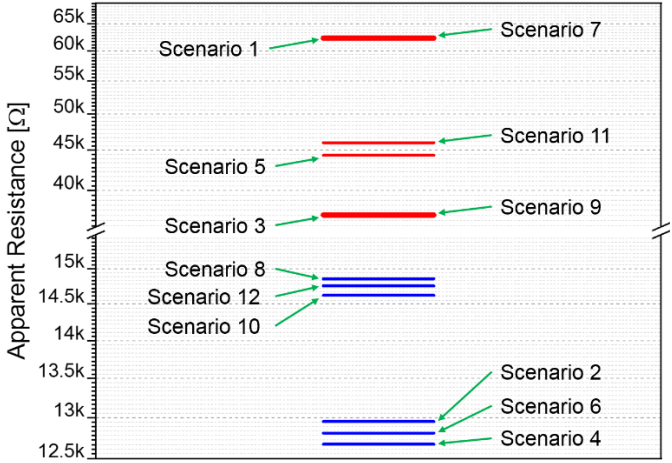

Fig. 6. The apparent resistances for a 100×100 (10kbit) resistive memory array using the 12 scenarios discussed before (red: high, blue: low). The chosen selector parameters are $\eta = 1.8$, $I_s = 10^{-12}A$. The calculations were done using the V/2 voltage scheme and a line resistance $R_l = 5\Omega$.

We concentrate mainly on simulating the common V/2 and V/3 read selection scheme. In the V/2 scheme, all the unselected WL and BL are biased at half the selected cell's read voltage. Hence, most of the leakage current is expected to be due to the half-selected cells. Those are the cells that share the BL with the selected cell, see Fig. 1. The V/3 scheme involves biasing all the unselected WL V/3, while all the unselected BL are biased at 2V/3. In a similar way to the V/2 scheme, the leakage current for the V/3 is primarily generated from all the partially selected cells that share BL with the selected cell.

III. RESULTS AND DISCUSSION

To demonstrate the reliability of the results obtained from this tool, Fig. 6 shows the general trend of the calculated apparent resistance and indicated the values obtained from every scenario. The results show that apparent resistance for the scenarios where $R_{select} = R_{high}$ are split into three groups corresponding to those when the unselected cells are at low, random and high resistance states. When $R_{unselect} = R_{high}$, the apparent resistance was calculated to be the largest. On the other hand, when $R_{unselect} = R_{low}$, the apparent resistance was calculated to be smallest. This is due to the large current contributed from each individual half-selected cell, i.e. the unselected cells that share the BL with the selected cell. The apparent resistance calculated when the cell is at a random state falls in between the extremes mentioned above. In each of the three groups where $R_{select} = R_{high}$, there are two scenarios that have close apparent resistances, corresponding to scenarios where the selected cell is located at the closest $(1,1)$ and furthest corners $(100, 100)$ to the voltage source and ground. I.e. the small difference in apparent resistance observed between scenarios 7 and 1, 11 and 5 and finally 9 and 3 is related to the effect of line resistance.

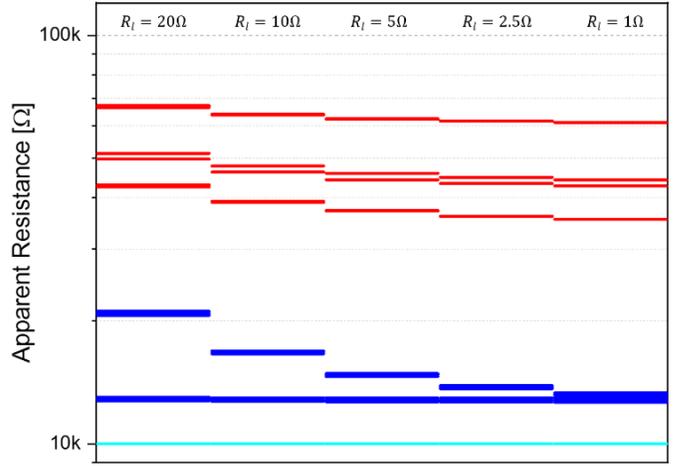

Fig. 7. The effect of increasing $R_l$ on the apparent resistance for a 100×100 1D1R array. The chosen selector parameters are $\eta = 1.8$, $I_s = 10^{-12}A$. The calculations were done using the V/2 voltage scheme.

The effect of line resistance is much more obvious for scenarios where $R_{select} = R_{low}$. Fig. 6 shows two groups of apparent resistances corresponding to the location of the selected cell. Scenarios 8, 12 and 10 that select cell $(100,100)$, show larger apparent resistance in comparison to that obtained from scenarios 2, 6 and 4, that select $(1,1)$, due to the effect of line resistance. The contribution of half-selected cells is much less influential when the selected cell has resistance $R_{low}$.

Reducing $R_l$ reduces the gap between the apparent resistance calculated for scenarios 8, 12 and 10 and scenarios 2, 6 and 4 as shown in Fig. 7. The line resistance was swept from 20 Ω to 1 Ω for a 100×100 1D1R array read using the V/2 scheme. The line resistance was found to have a much smaller effect on the apparent resistance calculated for scenarios where $R_{select} = R_{high}$. This is due to the large resistance ratio between the selected cell and line resistance when the former quals $R_{high}$ compared to that when it equals $R_{low}$. It is easy to notice that the apparent resistance calculated when $R_{select} = R_{high}$ is much smaller than its actual resistance. This is expected to be the case as the array size increases.

Increasing the array size increases the number of half-selected cells and the total current contributed from them, therefore reducing the apparent resistance. We simulated the 12 scenarios for four different array sizes to investigate that, and the results are plotted in Fig. 8. When the array size is changed from 10×10 to 200×200, the apparent resistance when the selected cell is high can reduce by an order of magnitude. Therefore, dramatically reducing the gap between the apparent resistance for high and low resistance state selected cells, hence reducing the sense margin. The low apparent resistance for $R_{high}$ selected cells can be increased for an array by optimizing the array's selector parameters, which will be investigated later.

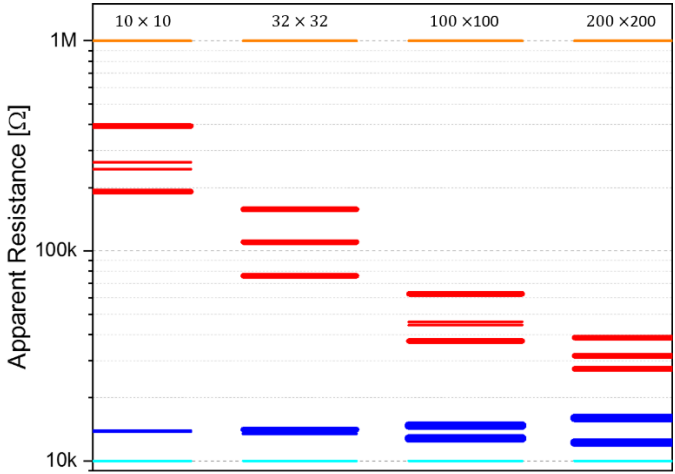

Fig. 8. The effect of changing array size on the apparent resistance calculated for a 100×100 resistive memory array using the 12 scenarios. The chosen selector parameters are $\eta = 1.8$, $I_s = 10^{-12} A$. The calculations were done using the V/2 voltage scheme and a line resistance $R_l = 5\Omega$. Cyan and orange indicate actual $R_{high}$ and $R_{low}$ respectively.

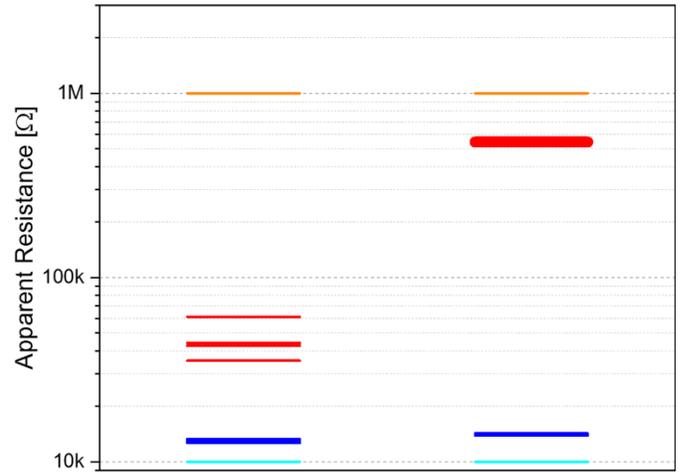

Fig. 9. The apparent resistances calculated for a 100×100 1D1R array using V/2 (left), and V/3 (right) voltage read schemes. The chosen selector parameters are $\eta = 1.8$, $I_s = 10^{-12} A$, while the line resistance $R_l = 1\Omega$. Cyan and orange indicate actual $R_{high}$ and $R_{low}$ respectively.

The choice of the voltage selection scheme can play a key role in determining the array's electrical behavior in reading operations. We used the tool to plot the apparent resistance calculated for the 12 scenarios using the V/2 and V/3 selection schemes. Fig. 9 shows a greater difference in the apparent resistance between scenarios where the selected cell is $R_{high}$ and $R_{low}$ for the V/3 scheme, compared to that for the V/2 scheme. This is because the current contributed by the partially selected cells that share the BL with the selected cell, is much smaller in the first scheme than in the latter, due to the smaller potential difference across those cells and the presence of the non-linear selector. The increase in resistance observed in those scenarios is caused due to smaller current leakage through unselected $R_{low}$ cells that share the BL with $R_{select}$. In addition, another feature of the V/3 scheme makes unselected cells, that constitute the vast majority of cells in the array, reversed biased. The reverse biasing nature of the V/3 select scheme makes it more suitable for rectifying diodes than for the V/2 scheme. The difference in the apparent resistance between scenarios where $R_{select} = R_{high}$ in the V/3 scheme is smaller than that for the V/2 scheme. This means that the state of unselected cells become less important for the V/3 selection scheme. This is expected to be the case due to the smaller potential difference across the partially selected cells in the V/3 scheme compared to the V/2 scheme, where their selectors play a key role in greatly reducing the current when the input voltage is reduced to V/3.

Ideally, a selector in a crossbar array switches on sharply at a voltage higher than V/2 (in the V/2 scheme) but lower than the read voltage V. However, the IV characteristics of a diode in series with a resistor follows eq. 2, plotted in Fig. 10, as the selector's reverse saturation current, $I_s$, and ideality factor, $\eta$, are changed. Therefore, optimizing the selector requires optimizing $I_s$ and $\eta$ to obtain the maximum sense margin from the V/3 and V/2 reading voltage schemes. We first plotted the sense margin as a function of $I_s$, as shown in Fig. 11. We found that for $I_s$ ranging from $10^{-14}$ to $10^{-10}$ the sense margin peaks around $I_s = 10^{-12} A$ at 7.1% for the V/3 scheme, while it peaks near $I_s = 10^{-13} A$ at 5.8% for the V/2 scheme. The sense margin reduces as $I_s$ is increased or decreased for both schemes. When $I_s$ is too large, the ratio of currents contributed by partially selected cells to that contributed by the fully selected cell will increase, therefore increasing the sense voltage read from scenario 9, where the selected cell is at a high resistance state, in comparison to scenario 8, hence reducing the sense margin. On the other hand, when $I_s$ is too small, the read voltage from scenario 8, where the selected cell is at low resistance state will reduce, hence reducing the sense margin. This also explains the similar behavior observed in Fig. 12. The simulations in the latter figure were done by fixing $I_s$ to the value at which the maximum sense margin was achieved in the previous figure while adjusting $\eta$. Increasing $\eta$ has a similar effect on the IV characteristics as that in decreasing $I_s$ as was demonstrated in Fig. 10, and vice versa. We found that the sense margin is optimized at $\eta = 1.5$ and $\eta = 1.7$ in the V/3 and V/2 schemes, respectively. However, the sense margin reduces as $\eta$ is deviated away from those values. In addition, according to eq. 2, changing T has a very similar effect to changing $\eta$. There are significant differences between the curves obtained from the V/3 and V/2 selection schemes. Firstly, the optimum sense margin for the V/3 scheme is larger than that for the V/2 scheme. Secondly, the sense margin optimum points along the $I_s$ and $\eta$ axis were different for the two schemes. Thirdly, the widths of the curves are also different for the two schemes, making the V/3 scheme more robust to fluctuations in the selector characteristics, that may arise due to fabrication or material non-uniformity problems. To better illustrate this behavior, we plot a 3D diagram showing the sense margin changing as a function of a range of different $I_s$ and $\eta$ for the V/3 selection scheme, see Fig. 13. Similar behavior can be shown using the selection scheme V/2, see Fig. 14.

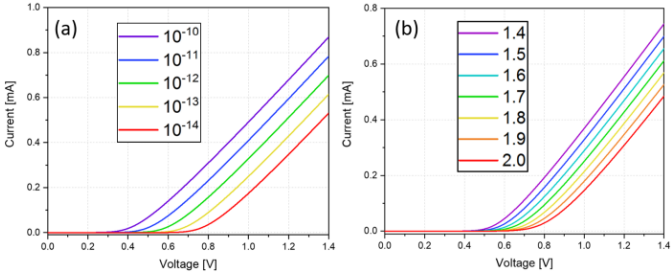

Fig. 10. The IV characteristics of a diode in series with a resistor plotted using Eq. 2 shows the shift in threshold voltage as $I_s$ (a) and $\eta$ (b) are changed. The plots were made for resistance R_low at temperature 300K.

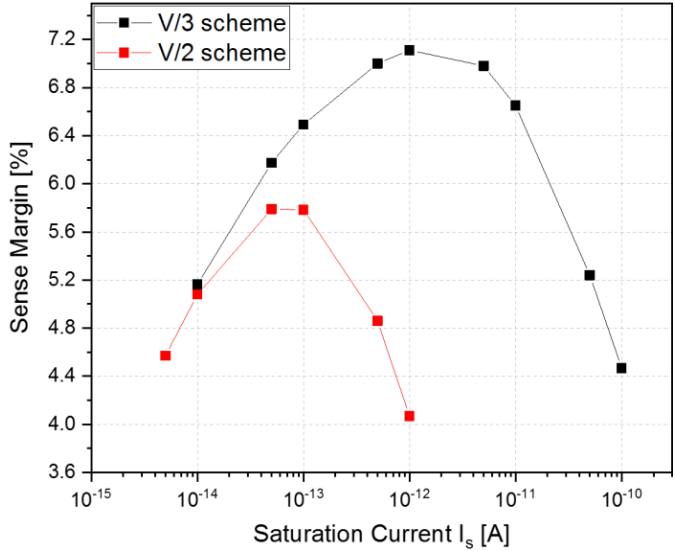

Fig. 11. The change in the sense margin as a function of $I_s$ for a 100×100 resistive memory array read using the V/3 and V/2 selection schemes. The figure shows a maximum sense margin of 7.1% achieved for $I_s = 10^{-12}\ A$. This reduces due to the effect of $I_s$ on the diode threshold voltage. In those calculation, $\eta = 1.7$ and $R_l = 1\ \Omega$ for both schemes.

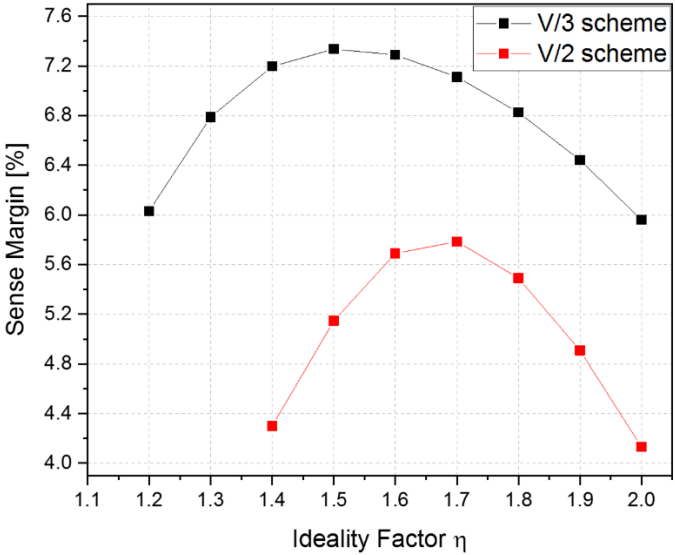

Fig. 12. The change in the sense margin as a function of $\eta$ for a 100×100 resistive memory array read using V/3 and V/2 selection schemes. The figure shows a maximum sense margin of 7.3% achieved for $\eta = 1.5$. This reduces due to the effect of $\eta$ on the diode threshold voltage. In those simulations, $I_s = 10^{-12}\ A$ and $I_s = 10^{-13}\ A$ for the V/3 and V/2 selection schemes, while $R_l = 1\ \Omega$ for both schemes.

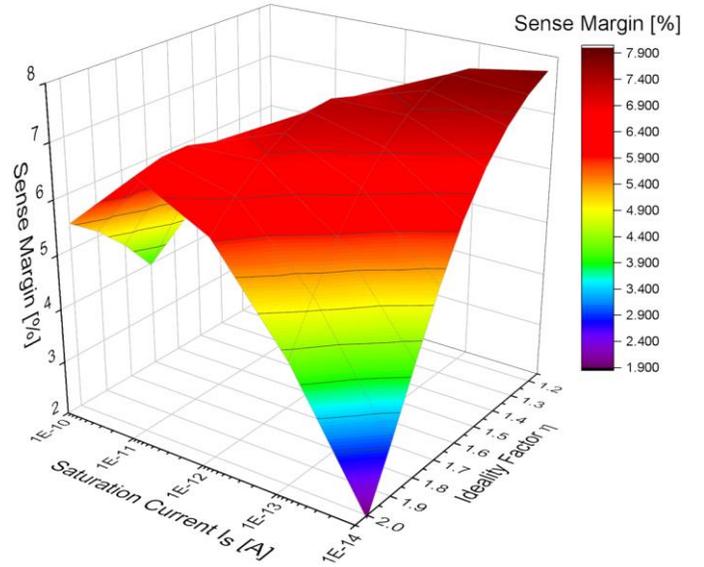

Fig. 13. A 3D plot showing the effect of changing $\eta$ and $I_s$ on the sense margin for a 100×100 resistive memory array read using the V/3 selection scheme.

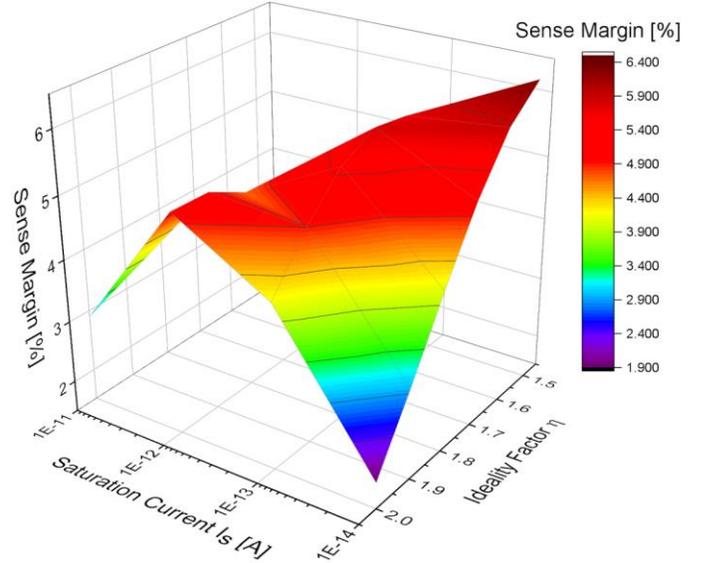

Fig. 14. A 3D plot showing the effect of changing $\eta$ and $I_s$ on the sense margin for a 100×100 resistive memory array read using the V/2 selection scheme.

In designing a crossbar memristor array, the dependence of $V_{sens\_BL}$ on $R_{sens}$ should be studied in order to optimize the sense margin. Choosing very large $R_{sens}$ reduces the sense margin due to increased RMS noise voltage ($v_{n\_RMS}$), otherwise known as Johnson-Nyquist noise. This can be calculated using Eq. 6

$$v_{n\_RMS} = \sqrt{4k_B T R_{sens} \Delta f} \qquad (6)$$

where $\Delta f$ is the operations bandwidth. For example, choosing $R_{sens} = 1\ M\Omega$, i.e. equivalent to $R_{high}$, $v_{n\_RMS}$ becomes higher than $4\ mV$, when operated at $T = 300\ K$ and $\Delta f = 1\ GHz$. This tool does not take Johnson-Nyquist noise effect on the sense margin, because it can be considered negligible for relatively small resistors. On the other hand, choosing $R_{sens}$ orders of magnitude smaller than $R_{low}$ reduces the sense margin due to the reduction in measured potential drop. This also increases the required sensitivity for performing voltage reads. In Fig. 15, we show the sense margin increases with increasing $R_{sens\_BL1}/R_{low}$ in a converging way. The result shows a larger sense margin obtained using the V/3 selection scheme, compared to the V/2 selection scheme as expected from the previous results.

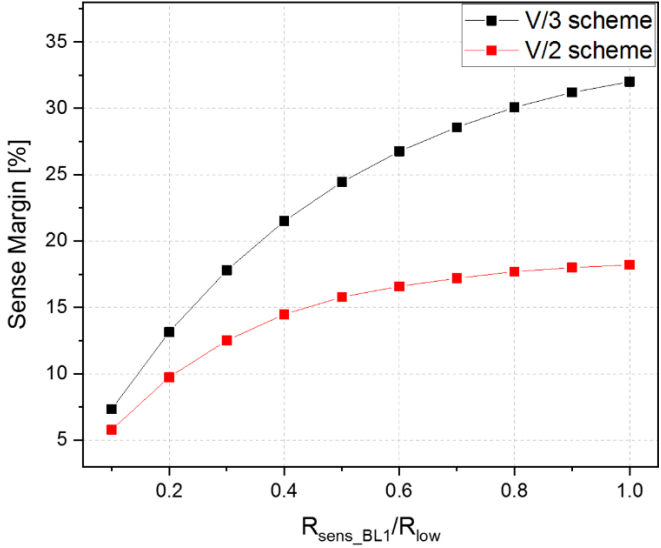

Fig. 15. A plot of worst-case scenario sensing margin as a function of sensing resistor for a $100 \times 100$ array using the optimised selector input parameters of $I_s = 10^{-12} A$ and $\eta = 1.5$ for the V/3 scheme and $I_s = 10^{-13} A$ and $\eta = 1.7$ for the V/2 selection scheme, $R_l = 1\,\Omega$ for both schemes.

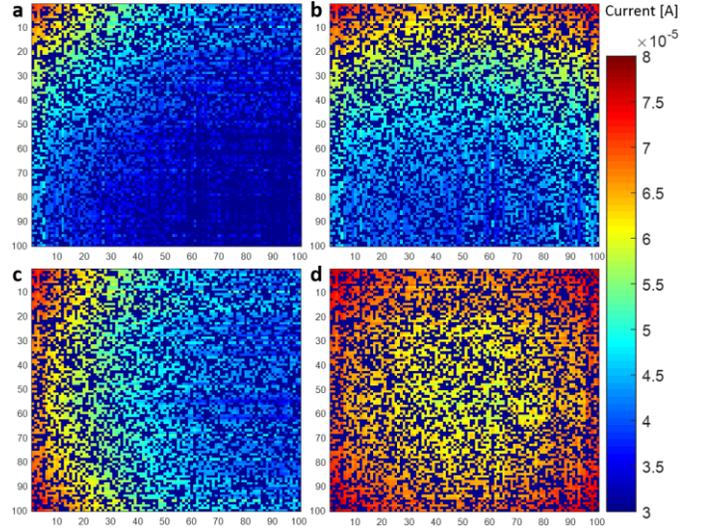

Fig. 16. A map of the read current for a 10kbit array at a random state. The maps show the effect of the non-zero line resistance $R_{WL} = R_{BL} = 1\Omega$ on the current dissipated at cells near the edges of the array. The four maps show the currents obtained for (a) single side biasing (left) and single side grounding (top), (b) double side biasing and single side grounding, (c) single side biasing and double side grounding and (d) finally double side biasing and double side grounding. The selector input parameters were chosen such that $\eta = 1.7$ and $I_s = 10^{-12} A$.

We then use the tool to plot the current dissipated through each cell in a 100x100 selector-memristor array at a random state. In this test, we compare the effect of biasing the array from both sides and from a single side of the WL on the overall current dissipated through the cells. Fig. 16 shows four different current maps. We start by a single side biasing, where the voltage source is located on the left-hand side of the array, such that $V_{app\_WL1}(i) = 1$, and the array is grounded through $R_{sens\_BL1}$. The sense resistors on the other sides of WL and BL were set to $R_{sens\_BL2}(i) = R_{sens\_WL2}(i) = 100\,M\Omega$. Fig. 16 (a) shows that the largest current dissipates at the top left corner at cell (1,1). Because this cell is located the closest to the voltage source and ground, therefore, it has no line resistance along its shortest current path. On the other hand, cell (100,100) has the maximum line resistance along its path contributed by both the WL ($R_{WL}$) and BL ($R_{BL}$). Fig. 16 (b) shows the same array biased from both sides of the WL, such that $V_{app\_WL1}(i) = V_{app\_WL2}(i) = 1$ and $R_{sens\_WL2}(i) = 10\,\Omega$. With a single side grounding, the current reduces for cells further away from the ground side. This is caused by larger line resistance along the path of cells furthest away from the grounding side. In this arrangement, between the low resistance state cells, the cell that dissipates the least current should be (100,50), because it is the furthest from both sides of the dual side biased WL. Similarly, when the array is biased from a single side and grounded from both sides, the cell that dissipates the least current is (50,100), as shown in Fig. 16 (c). The current dissipated in cases (b) and (c) is quantitatively symmetrical in this simulation, except that the trend is rotated by 90 degrees, however, this is only the case because $R_{WL} = R_{BL}$ in those simulations. With dual side biasing and dual side grounding, the ratio between the largest and smallest current dissipated by a cell is reduced to 1.36 compared to 2.4 when the single side bias and ground case is applied. In other words, dual biasing and grounding can help to improve the current dissipation uniformity throughout the cell, especially for arrays that suffer from relatively high $R_l$ values. The current is much lower for high resistance state cells as can be demonstrated by the navy colored points showing currents off the scale's minimum. A similar plot can be made using this tool to demonstrate the effect of the finite line resistance on the node voltages $V_{WL}$ and $V_{BL}$ and the power dissipated in every cell.

IV. DISCUSSION AND SUMMARY

To give a brief indication of the processing time this tool takes, the 1D1R selector-memristor tool was used to simulate a 1000×1000 array using a PC with 6 core, 12 threaded Intel Core i7 running at 3.5GHz with a 40GB DDR3 memory. In this setup, a single array state simulation took approximately 30 minutes. A single simulation of a 100×100 array typically takes 30 s to 5 minutes on the same PC, depending on the number of iterations needed to achieve the required accuracy. The processing time needed to simulate larger matrices is expected to increase linearly, due to the efficiency in using processor's time, however, practical simulation of many Mb arrays may become only possible through supercomputer clusters.

The tool can also be used to simulate arrays of different shapes based on the input resistance matrix. While in this work the WL and BL resistance had a single value, the code can be adjusted to include a matrix of WL and BL resistance to accurately simulate real memristive arrays.

The choice behind the values of $R_{low}$ and $R_{high}$ used in these simulations was motivated by other research works demonstrating that those values are within the expected resistance range of high and low resistance states for phase-change memory devices. An example is those based on electroplated GeSbTe (GST), which triggered our work in this field [13], [23], [24]. Based on the optimized selector parameters we have proposed in this work, there are different materials that can be used to successfully realize many kbits arrays with an achievable sense margin beyond 30%. For example, silicon diodes are known to have $\eta$ between 1 and 2 and $I_s$ in the range of $10^{-12}A$ [25]. However, material compatible selectors to GST such as those based on chalcogenides, can also be proposed. There have been several research works demonstrating ovonic threshold switching as a technology for realizing selectors for resistive and phase change memory arrays. The electrical characteristics of GeSe ovonic threshold switches were shown to be highly tunable with doping material and concentration, which makes them versatile for GST phase change memory applications [26]–[30]. In addition, $GeTe_6$ was shown to have excellent electrical properties, but may not meet the thermal stability needed [31]–[35]. This is also subject to the memristor and selector device dimensions which is expected to be very small for commercialized technologies. However, we anticipate that our tool is highly flexible to keep up with simulating memristor arrays as the field progresses and different device electrical properties are explored.

In summary, we developed a MATLAB-based tool that allows performing electronic analysis of 1D1R arrays. The tool demonstrated the important effects of line resistance, voltage array size, selection scheme and selector's ideality factor and reverse saturation current on the successful design of memristor arrays. The work explored different array state scenarios to investigate the contribution of sneak paths on the apparent resistance recorded for 100×100 memristor array as the line resistance, array size, and bias scheme are changed. We demonstrated the expected behavior of sense margin as the selectors' ideality factor and reverse saturation current is changed for the V/2 and V/3 biasing schemes for GST phase-change memristors combined with silicon diodes or ovonic threshold switching selectors. Finally, 2D maps were plotted to show the importance of correct biasing and grounding of an array on the current distribution uniformity throughout the array.

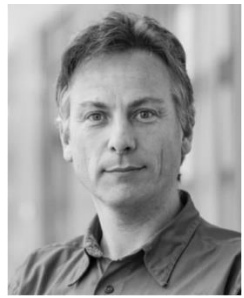
**C. H. Kees de Groot** received the master's degree in physics from the University of Groningen, Groningen, the Netherlands, in 1994, and the Ph.D. degree from the University of Amsterdam in 1998 for research carried out at the Philips Research laboratories in Eindhoven, the Netherlands. Subsequently, he was a Research Fellow with the Massachusetts Institute of Technology, Cambridge, MA, USA, where he conducted research on spin tunnel junction and phase change materials. Since 2000, he has been with the Department of Electronics and Computer Science, University of Southampton, Southampton, U.K., where he has been a Full Professor since 2012. He is an Author of more than 125 journal publications. His main interest is the integration of novel nano-materials and devices with silicon electronics processing with particular emphasis on the semiconducting and dielectric properties of oxides, chalcogenides, and carbides. His recent breakthroughs in these areas includes the first 100-nm GeSbTe phase change memory by non-aqueous electrodeposition, SiC resistive memory with a record nine orders of magnitude on/off ratio, and functional oxide nanostructures resulting in plasmonic devices with ultrafast modulation of optical and dielectric properties of ITO, AZO, and VO2 by optical, electrical, and thermal methods.

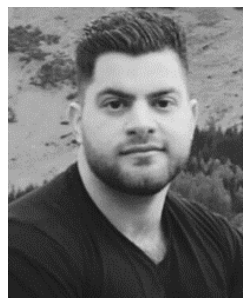
**Yasir J Noori** received his Bachelor of Engineering with first-class honors in Electronic Engineering and Physics from the University of Dundee in 2013. He then received a prestigious Faculty of Science and Technology studentship to study his PhD in Quantum Key Distribution from Lancaster University in 2017. He has over 15 publications, 2 book chapters, one patent and made many conference contributions. Yasir is currently a Research Fellow-Nanotechnology Expert within a £6.3m EPSRC programme grant for making Advanced Devices by ELectroPlaTing (ADEPT) where he develops phase change memory and infrared detectors using materials grown by electroplating. Yasir is interested in integrated photonics, phase change materials and quantum information. He uses his expertise in device fabrication and e-beam lithography for applications in phase change memories, quantum technologies and photonic circuits. He is a member of the Institute of Electrical and Electronic Engineering (IEEE) and the Institute of Physics (IOP)